\documentclass{svproc}
\usepackage{url}
\usepackage{graphicx}
\usepackage{ccicons}                              

\newcommand{\RNum}[1]{%
  \textup{\uppercase\expandafter{\romannumeral#1}}%
}

\newcommand{\qu}[1]{``#1''}
\usepackage{blindtext}
\usepackage{graphicx}
\usepackage[T1]{fontenc}
\usepackage{amssymb}
\usepackage{graphicx}
\usepackage{amssymb}
\usepackage{tabularx}
\usepackage{ragged2e}
\usepackage{comment}
\usepackage{comment}
\usepackage[utf8]{inputenc}
\usepackage[T1]{fontenc}
\usepackage{lmodern} 
\usepackage{diagbox}

\usepackage{amsmath}
\usepackage{lscape} 
\usepackage[figuresright]{rotating}
\usepackage{subcaption}
\usepackage{multirow}
\usepackage{amsmath}
\usepackage[colorlinks=true,
            linkcolor=red,
            urlcolor=blue,
            citecolor=blue]{hyperref}
\usepackage{comment}

\begin{document}
\mainmatter              
\title{Women in ISIS Propaganda: A Natural Language Processing Analysis of Topics and Emotions in a Comparison with Mainstream Religious Group}
\titlerunning{Women in ISIS Propaganda: A Natural Language Processing Analysis}
%
%

\author{Mojtaba Heidarysafa\inst{1} \and Kamran Kowsari\inst{1} \and Tolu Odukoya\inst{2} \and Philip Potter \inst{3} \and  Laura E. Barnes\inst{1,3} \and Donald E. Brown \inst{1,3}}
\authorrunning{M. Heidarysafa et al.} 
%
%
\institute{Department  of  Systems \&  Information  Engineering,  University  of Virginia, Charlottesville, VA, USA
\and
School  of  Data  Science,  University  of  Virginia,  Charlottesville,  VA, USA
\and 
Department of Politics, University of Virginia, Charlottesville, VA, USA
\\~\\
$^*$ Co-corresponding authors:~\href{mailto:mh4pk@virginia.edu}{mh4pk@virginia.edu}}

\maketitle              

\begin{abstract}
Online propaganda is central to the recruitment strategies of extremist groups and in recent years these efforts have increasingly extended to women. To investigate ISIS' approach to targeting women in their online propaganda and uncover implications for counterterrorism, we rely on text mining and natural language processing (NLP). Specifically, we extract articles published in Dabiq and Rumiyah (ISIS's online English language publications) to identify prominent topics. To identify similarities or differences between these texts and those produced by non-violent religious groups, we extend the analysis to articles from a Catholic forum dedicated to women. We also perform an emotional analysis of both of these resources to better understand the emotional components of propaganda. We rely on Depechemood~(a lexical-base emotion analysis method) to detect emotions most likely to be evoked in readers of these materials. The findings indicate that the emotional appeal of ISIS and Catholic materials are similar.
\end{abstract}

\section{Introduction}
Since its rise in 2013, the Islamic State of Iraq and Syria~(ISIS) has utilized the Internet to spread its ideology, radicalize individuals, and recruit them to their cause. In comparison to other Islamic extremist groups,  ISIS' use of technology was more sophisticated, voluminous, and targeted. For example, during ISIS' advance toward Mosul, ISIS related accounts tweeted some 40,000 tweets in one day ~\cite{farwell2014media}.However, this heavy engagement forced social media platforms to institute policies to prevent unchecked dissemination of terrorist propaganda to their users, forcing ISIS to adapt to other means to reach their target audience. 

One such approach was the publication of online magazines in different languages including English. Although discontinued now, these online resources provided a window into  ISIS ideology, recruitment, and how they wanted the world to perceive them. For example, after predominantly recruiting men, ISIS began to also include articles in their magazines that specifically addressed women. ISIS encouraged women to join the group by either traveling to the caliphate or by carrying out domestic attacks on behalf of ISIS in their respective countries. This tactical change concerned both practitioners and researchers in the counterterrorism community. New advancements in data science can shed light on exactly how the targeting of women in extremist propaganda works and whether it differs significantly from mainstream religious rhetoric. 

We utilize natural language processing methods to answer three questions:
\begin{itemize}
    \item What are the main topics in women-related articles in ISIS' online magazines?
    \item What similarities and/or differences do these topics have with non-violent, non-Islamic religious material addressed specifically to women?
    \item What kind of emotions do these articles evoke in their readers and are there similarities in the emotions evoked from both ISIS and non-violent religious materials?
\end{itemize}
As these questions suggest, to understand what, if anything, makes extremist appeals distinctive, we need a point of comparison in terms of the outreach efforts to women from a mainstream, non-violent religious group. For this purpose, we rely on an online Catholic women's forum. Comparison between Catholic material and the content of ISIS' online magazines allows for novel insight into the distinctiveness of extremist rhetoric when targeted towards the female population. To accomplish this task, we employ topic modeling and an unsupervised emotion detection method.

The rest of the paper is organized as follows: in Section~\ref{Sec2}, we review related works on ISIS propaganda and applications of natural language methods. Section~\ref{sec:Data} describes data collection and pre-processing. Section~\ref{Sec3} describes in detail the approach.  Section~\ref{Sec4} reports the results, and finally, Section~\ref{Sec5} presents the conclusion.

\section{Related Work}\label{Sec2}
Soon after ISIS emerged and declared its caliphate, counterterrorism practitioners and political science researchers started to turn their attention towards understanding how the group operated. Researchers investigated the origins of ISIS, its leadership, funding, and how they rose became a globally dominant non-state actor~\cite{laub2014islamic}. This interest in the organization's distinctiveness immediately led to inquiries into ISIS' rhetoric, particularly their use of social media and online resources in recruitment and ideological dissemination. For example, Al-Tamimi examines how ISIS differentiated itself from other jihadist movements by using social media with unprecedented efficiency to improve its image with locals~\cite{al2014dawn}. One of ISIS' most impressive applications of its online prowess was in the recruitment process. The organization has used a variety of materials, especially videos, to recruit both foreign and local fighters. Research shows that ISIS propaganda is designed to portray the organization as a provider of justice, governance, and development in a fashion that resonates with young westerners~\cite{gates2015social}. This propaganda machine has become a significant area of research, with scholars such as Winter identifying key themes in it such as brutality, mercy, victimhood, war, belonging and utopianism.~\cite{winter2015virtual}. However, there has been insufficient attention focused on how these approaches have particularly targeted and impacted women. This is significant given that scholars have identified the distinctiveness of this population when it comes to nearly all facets of terrorism. 

ISIS used different types of media to propagate its messages, such as videos, images, texts, and even music. Twitter was  particularly effective and the Arabic Twitter app allowed ISIS to tweet extensively without triggering spam-detection mechanisms the platform uses~\cite{farwell2014media}. Scholars followed the resulting trove of data and this became the preeminent way in which they assess ISIS messages. For example, in ~\cite{bodine2016examining} they use both lexical analysis of tweets as well as social network analysis to examine ISIS support or opposition on Twitter. Other researchers used data mining techniques to detect pro-ISIS user divergence behavior at various points in time~\cite{rowe2016mining}. By looking at these works, the impact of using text mining and lexical analysis to address important questions becomes obvious. Proper usage of these tools allows the research community to analyze big chunks of unstructured data. This approach, however, became less productive as the social media networks began cracking down and ISIS recruiters moved off of them. 

With their ability to operate freely on social media now curtailed, ISIS recruiters and propagandists increased their attentiveness to another longstanding tool--English language online magazines targeting western audiences. Al Hayat, the media wing of ISIS, published multiple online magazines in different languages including English. The English online magazine of ISIS was named Dabiq and first appeared on the dark web on July 2014 and continued publishing for 15 issues. This publication was followed by Rumiyah which produced 13 English language issues through September 2017. The content of these magazines provides a valuable but underutilized resource for understanding ISIS strategies and how they appeal to recruits, specifically English-speaking audiences. They also provide a way to compare ISIS' approach with other radical groups. Ingram compared Dabiq contents with Inspire (Al Qaeda  publication) and suggested that Al Qaeda heavily emphasized identity-choice, while ISIS' messages were more balanced between identity-choice and rational-choice~\cite{ingram2017analysis}. In another research paper, Wignell et al.~\cite{wignell2017mixed} compared Dabiq and Rumiah by examining their style and what both magazine messages emphasized. Despite the volume of research on these magazines, only a few researchers used lexical analysis and mostly relied on experts' opinions.~\cite{vergani2015evolution} is one exception to this approach where they used word frequency on 11 issues of Dabiq publications and compared attributes such as anger, anxiety, power, motive, etc.

This paper seeks to establish how ISIS specifically tailored propaganda targeting western women, who became a particular target for the organization as the ``caliphate'' expanded. Although the number of recruits is unknown, in 2015 it was estimated that around 10 percent of all western recruits were female~\cite{perevsin2015fatal}. Some researchers have attempted to understand how ISIS propaganda targets women. Kneip, for example, analyzed women's desire to join as a form of emancipation~\cite{kneip2016female}. We extend that line of inquiry by leveraging technology to answer key outstanding questions about the targeting of women in ISIS propaganda. 

To further assess how ISIS propaganda might affect women, we used emotion detection methods on these texts. Emotion detection techniques are mostly divided into lexicon-base or machine learning-base methods. Lexicon-base methods rely on several lexicons while machine learning (ML) methods use algorithm to detect the elation of texts as inputs and emotions as the target, usually trained on a large corpus. Unsupervised methods usually use Non-negative matrix factorization~(NMF) and Latent Semantic Analysis~(LSA)~\cite{canales2014emotion} approaches. An important distinction that should be made when using text for emotion detection is that emotion detected in the text and the emotion evoked in the reader of that text might differ. In the case of propaganda, it is more desirable to detect possible emotions that will be evoked in a hypothetical reader. In the next section, we describe methods to analyze content and technique to find evoked emotions in a potential reader using available natural language processing tools.

\section{Data Collection \& Pre-Processing}\label{sec:Data}
\subsection{Data collection}
Finding useful collections of texts where ISIS targets women is a challenging task. Most of the available material are not reflecting ISIS' official point of view or they do not talk specifically about women. However, ISIS' online magazines are valuable resources for understanding how the organization attempts to appeal to western audiences, particularly women. Looking through both Dabiq and Rumiyah, many issues of the magazines contain articles specifically addressing women, usually with `` to our sisters '' incorporated into the title. Seven out of fifteen Dabiq issues and all thirteen issues of Rumiyah contain articles targeting women, clearly suggesting an increase in attention to women over time. 

We converted all the ISIS magazines to texts using pdf readers and all articles that addressed women in both magazines (20 articles) were selected for our analysis.
To facilitate comparison with a mainstream, non-violent religious group, we collected articles from catholicwomensforum.org,  an online resource catering to Catholic women. We scrapped 132 articles from this domain. While this number is large, the articles themselves are much shorter than those published by ISIS. 
These texts were pre-processed by tokenizing the sentences and eliminating non-word tokens and punctuation marks. Also, all words turned into lower case and numbers and English stop words such as ``our, is, did, can, etc. '' have been removed from the produced tokens. For the emotion analysis part, we used a spacy library as part of speech tagging to identify the exact role of words in the sentence. A word and its role have been used to look for emotional values of that word in the same role in the sentence.  

\subsection{Pre-Processing}
\subsubsection{Text Cleaning and Pre-processing}
Most text and document datasets contain many unnecessary words such as stopwords, misspelling, slang, etc.
In many algorithms, especially statistical and probabilistic learning algorithms, noise and unnecessary features can have adverse effects on system performance. In this section, we briefly explain some techniques and methods for text cleaning and pre-processing text datasets~\cite{kowsari2019text}.  

\subsubsection{Tokenization}
Tokenization is a pre-processing method which breaks a stream of text into words, phrases, symbols, or other meaningful elements called tokens~\cite{verma2014tokenization}.  The main goal of this step is to investigate the words in a sentence~\cite{verma2014tokenization}. Both text classification and text mining requires a parser which processes the tokenization of the documents; for example:\\
sentence~\cite{aggarwal2018machine} :

\textit{After sleeping for four hours, he decided to sleep for another four.}\\
In this case, the tokens are as follows:

\{\textit{\qu{After} \qu{sleeping} \qu{for} \qu{four} \qu{hours} \qu{he} \qu{decided} \qu{to} \qu{sleep} \qu{for} \qu{another} \qu{four}}\}.

\subsubsection{Stop words}
Text and document classification includes many words which do not hold important significance to be used in classification algorithms such as \{\textit{\qu{a},
\qu{about},
\qu{above},
\qu{across},
\qu{after},
\qu{afterwards},
\qu{again},$\hdots$}\}. The most common technique to deal with these words is to remove them from the texts and documents~\cite{saif2014stopwords}.

\subsubsection{Term Frequency-Inverse Document Frequency}

\textit{K Sparck Jones}~\cite{sparck1972statistical} proposed inverse document frequency~(IDF) as a method to be used in conjunction with term frequency in order to lessen the effect of implicitly common words in the corpus. 
IDF assigns a higher weight to words with either high frequency or low frequency term in the document.
This combination of TF and IDF is well known as term frequency-inverse document frequency (tf-idf). The mathematical representation of the weight of a term in a document by tf-idf is given in Equation~\ref{tf-idf}.

\begin{equation}\label{tf-idf}
    W(d,t)=TF(d,t)* log(\frac{N}{df(t)})
\end{equation}

Here N is the number of documents and~$df(t)$ is the number of documents containing the term t in the corpus. The first term in Equation~\ref{tf-idf} improves the recall while the second term improves the precision of the word embedding~\cite{tokunaga1994text}. Although tf-idf tries to overcome the problem of common terms in the document, it still suffers from some other descriptive limitations. Namely, tf-idf cannot account for the similarity between the words in the document since each word is independently presented as an index. 
However, with the development of more complex models in recent years, new methods, such as word embedding, have been presented that can incorporate concepts such as similarity of words and part of speech tagging.

\section{Method}\label{Sec3}
In this section, we describe our methods used for comparing topics and evoked emotions in both ISIS and non-violent religious materials.

\subsection{Content Analysis}
The key task in comparing ISIS material with that of a non-violent group involves analyzing the content of these two corpora to identify the topics. For our analysis, we considered a simple uni-gram model where each word is considered as a single unit. Understanding what words appear most frequently provides a simple metric for comparison. To do so we normalized the count of words with the number of words in each corpora to account for the size of each corpus. It should be noted, however, that a drawback of word frequencies is that there might be some dominant words that will overcome all the other contents without conveying much information.

\begin{figure}[t]
\centering
\includegraphics[width=0.65\columnwidth]{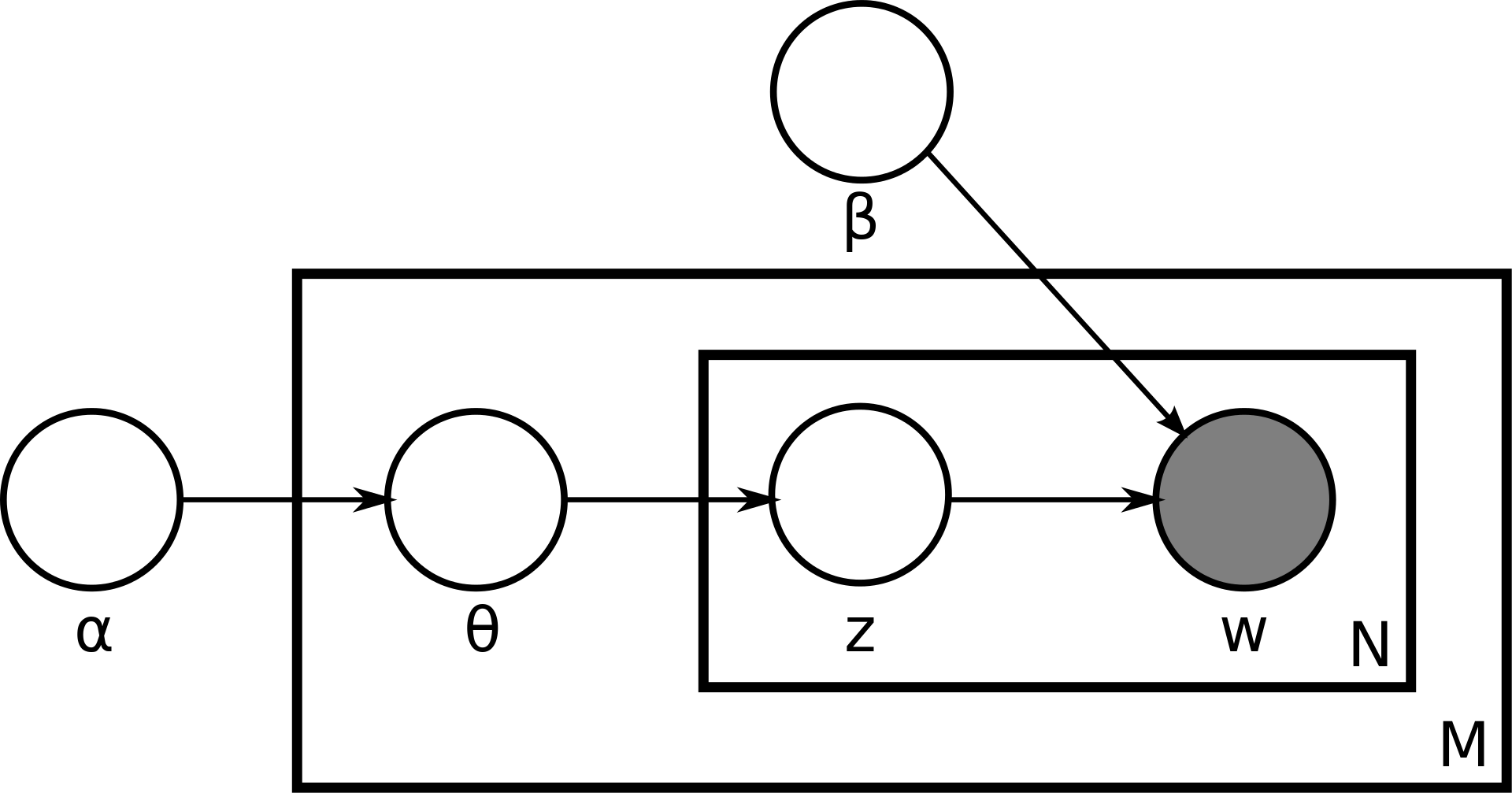}
\caption{Plate notation of LDA model}\label{fig:LDA}
\end{figure}

Topic modeling methods are the more powerful technique for understanding the contents of a corpus. These methods try to discover abstract topics in a corpus and reveal hidden semantic structures in a collection of documents. The most popular topic modeling methods use probabilistic approaches such as probabilistic latent semantic analysis~(PLSA) and latent Dirichlet allocation~(LDA).  LDA is a generalization of pLSA where documents are considered as a mixture of topics and the distribution of topics is governed by a Dirichlet prior~($\alpha$). Figure~\ref{fig:LDA} shows plate notation of general LDA structure where~$\beta$ represents prior of word distribution per topic and~$\theta$ refers to topics distribution for documents~\cite{blei2003latent}. Since LDA is among the most widely utilized algorithms for topic modeling, we applied it to our data. However, the coherence of the topics produced by LDA is poorer than expected. 

\begin{figure}[b]
\centering
\includegraphics[width=0.85\columnwidth]{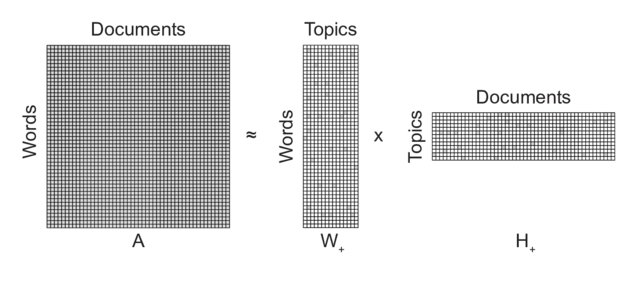}
\caption{NMF decomposition of document-term matrix~\cite{kuang2017crime}}\label{fig:nmf}
\end{figure}

To address this lack of coherence, we applied non-negative matrix factorization~(NMF). This method decomposes the term-document matrix into two non-negative matrices as shown in Figure~\ref{fig:nmf}. The resulting non-negative matrices are such that their product closely approximate the original data. Mathematically speaking, given an input matrix of document-terms $V$, NMF finds two matrices by solving the following equation~\cite{lee1999learning}:
\begin{equation*}
    \min_{W,H}\left\lVert V- WH\right\rVert_{F}\quad s.t \quad H \geq 0, W\geq 0.
\end{equation*}
Where W is topic-word matrix and H represents topic-document matrix.

NMF appears to provide more coherent topic on specific corpora. O'Callaghan et al. compared LDA with NMF and concluded that NMF performs better in corporas with specific and non-mainstream areas~\cite{o2015analysis}. Our findings align with this assessment and thus our comparison of topics is based on NMF. 
\subsection{Emotion detection}\label{emodetect}
Propaganda effectiveness hinges on the emotions that it elicits.  But detecting emotion in text requires that two essential challenges are overcome. 

\begin{figure}
\centering
\includegraphics[width=0.54\columnwidth]{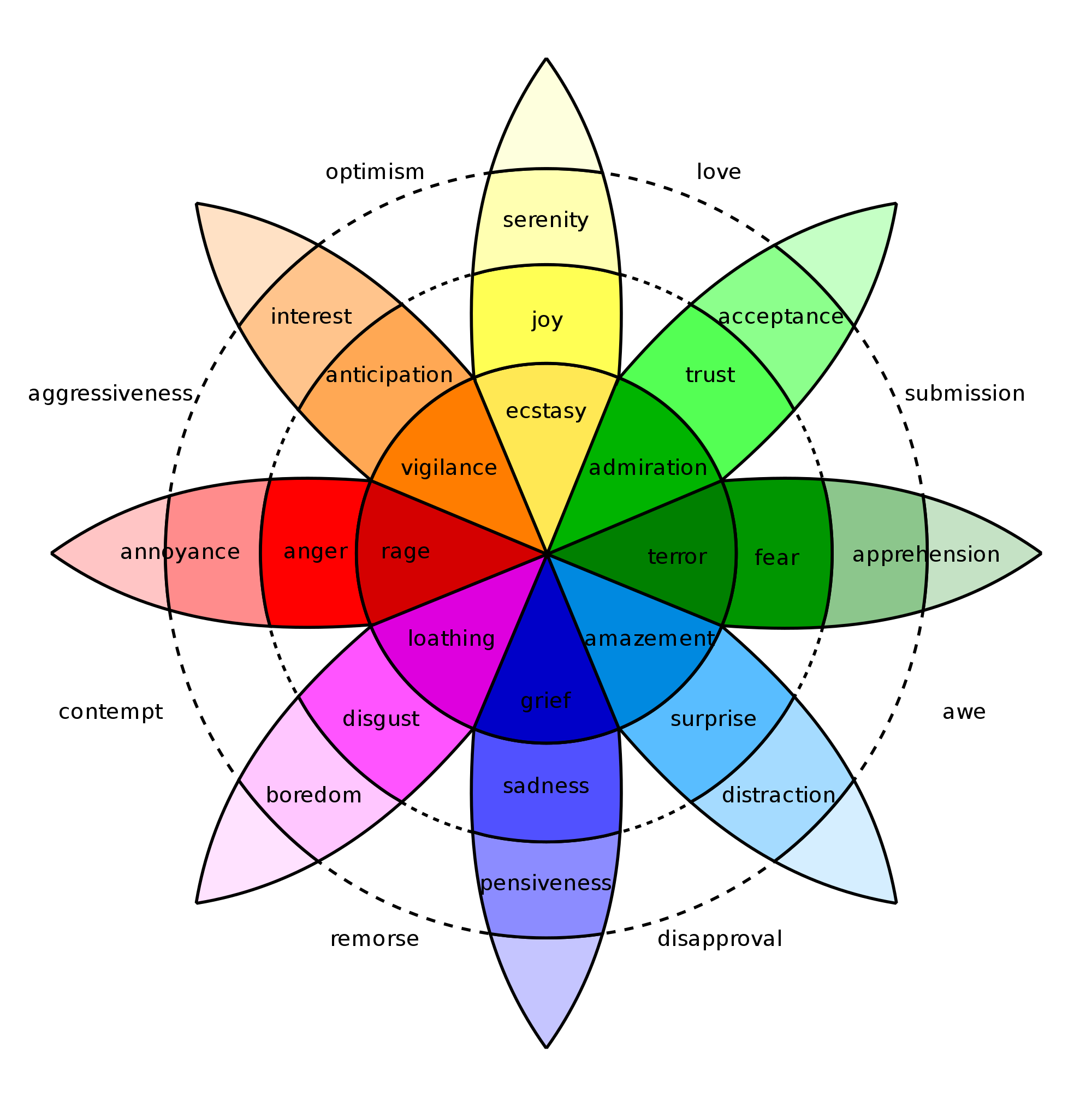}
\caption{2D representation of Plutchik wheel of emotions~\cite{plutchik1980general}}\label{fig:wheel}
\end{figure}

First, emotions are generally complex and emotional representation models are correspondingly contested. Despite this, some models proposed by psychologists have gained wide-spread usage that extends to text-emotion analysis. Robert Plutchik presented a model that arranged emotions from basic to complex in a circumplex as shown in Figure \ref{fig:wheel}. The model categorizes emotions into 8 main subsets and with addition of intensity and interactions it will classify emotions into 24 classes~\cite{plutchik2001nature}. Other models have been developed to capture all emotions by defining a 3-dimensional model of pleasure, arousal, and dominance. 

The second challenge lies in using text for detecting emotion evoked in a potential reader. Common approaches use either lexicon-base methods~(such as keyword-based or ontology-based model) or machine learning-base models~(usually using large corpus with labeled emotions)~\cite{canales2014emotion}. These methods are suited to addressing the emotion that exist in the text, but in the case of propaganda we are more interested in emotions that are elicited in the reader of such materials. The closest analogy to this problem can be found in research that seek to model feelings of people after reading a news article. One solution for this type of problem is to use an approach called Depechemood.

Depechemood is a lexicon-based emotion detection method gathered from crowd-annotated news~\cite{staiano2014depechemood}. Drawing on approximately 23.5K documents with average of 500 words per document from rappler.com, researchers asked subjects to report their emotions after reading each article. They then multiplied the document-emotion matrix and word-document matrix to derive emotion-word matrix for these words. Due to limitations of their experiment setup, the emotion categories that they present does not exactly match the emotions from the Plutchik wheel categories. However, they still provide a good sense of the general feeling of an individual after reading an article. The emotion categories of Depechemood are:
AFRAID, AMUSED, ANGRY, ANNOYED, DON'T CARE, HAPPY, INSPIRED, SAD. Depechemood simply creates dictionaries of words where each word has scores between 0 and 1 for all of these 8 emotion categories. We present our finding using this approach in the result section.

\begin{figure}[h]
\centering
\includegraphics[width=0.7\textwidth]{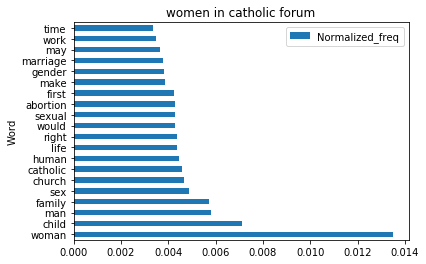}
\caption{Word frequency of most common words in catholic corpora}
\end{figure}

\begin{figure}[h]
\centering
\includegraphics[width=0.7\textwidth]{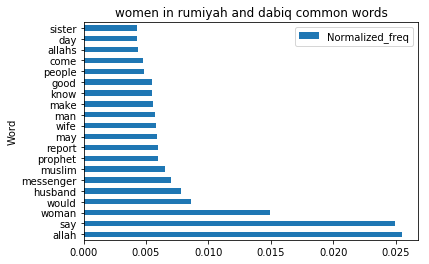}
\caption{Word frequency of most common words in dabiq corpora} \label{fig:word_count}
\end{figure}

\begin{table}[]
\centering
\caption{NMF Topics of women in ISIS}\label{table:isis}
\rotatebox{90}{
\begin{tabular}{l c c c c c c c c c c }
\hline
&\textbf{\begin{tabular}[c]{@{}c@{}}early islam \\ women\end{tabular}} & \textbf{\begin{tabular}[c]{@{}c@{}}Islam / \\ khilafah\end{tabular}} & \textbf{marriage}   & \textbf{islamic praying} & \textbf{women's life} & \textbf{hijrah}    & \textbf{islamic}   & \textbf{divorce}  & \textbf{motherhood} & \textbf{\begin{tabular}[c]{@{}c@{}}spousal \\ relationship\end{tabular}} \\ \hline
& \textbf{Topic 0}           & \textbf{Topic 1}          & \textbf{Topic 2}    & \textbf{Topic 3}         & \textbf{Topic 4}      & \textbf{Topic 5}   & \textbf{Topic 6}   & \textbf{Topic 7}  & \textbf{Topic 8} &   \textbf{ Topic 9}               \\ \hline
1 &  said              & wa               & husband    & masjid          & worldly      & islamic   & wala      & mourning & said       & spouses              \\ \hline
2 &  khadijah          & hijrah           & woman      & prayer          & spend        & state     & said      & iddah    & ibn        & said                 \\ \hline
3 &  munafiqin         & sallam           & sharah     & said            & regards      & khilafah  & bara      & widow    & jihad      & husband              \\ \hline
4 &  iman              & alayhi           & said       & masajid         & wearing      & abu       & sake      & home     & reported   & backbiting           \\ \hline
5 &  abu               & sallallhu        & wives      & home            & mat          & brothers  & enmity    & husband  & children   & listening            \\ \hline
6 &  ibn               & khilfah          & wife       & going           & menstruation & arrived   & kab       & ibn      & charity    & wife                 \\ \hline
7 &  dunya             & ab               & marry      & woman           & prophet      & knew      & salam     & said     & child      & abu                  \\ \hline
8 &  mother            & sisters          & ibn        & ibn             & life         & mujahidin & remained  & perfume  & abu        & divorce              \\ \hline
9 &  people            & radiyallhu       & married    & prophet         & clothing     & previous  & prayer    & woman    & wealth     & word                 \\ \hline
10 & asma              & ibn              & jihd       & hadith          & small        & soon      & shariah   & away     & flock      & problems             \\ \hline
11 & husband           & qurn             & duny       & reported        & lived        & children  & muslims   & night    & muslim     & instead              \\ \hline
12 & bakr              & islamic          & say        & prevent         & time         & later     & affection & wear     & cause      & backbite             \\ \hline
13 & believers         & praise           & permitted  & men             & family       & informed  & islam     & widows   & albukhari  & relationship         \\ \hline
14 & hearts            & state            & lawful     & default         & aishah       & hijrah    & anger     & house    & prophet    & tongue               \\ \hline
15 & killed            & said             & prohibited & muslim          & garment      & shariah   & aqidah    & sleep    & policy     & woman                \\ \hline
16 & know              & slavegirl        & lord       & leaving         & despite      & dua       & good      & married  & shepherd   & person               \\ \hline
17 & steadfast         & land             & islam      & abu             & follow       & journey   & religion  & clothing & soul       & brother              \\ \hline
18 & sumayyah          & firm             & fear       & stay            & concern      & returned  & return    & ab       & waging     & home                 \\ \hline
19 & sisters           & people           & man        & wives           & living       & umm       & state     & pregnant & lord       & hind                 \\ \hline
20 & spreading         & lands            & sister     & shariah         & food         & leave     & relatives & husbands & mother     & narrated             \\ \hline
\end{tabular}
}
\end{table}

\begin{table}[]
\centering
\caption{NMF Topics of women in catholic forum}\label{table:cat}
\rotatebox{90}{
\begin{tabular}{l c c c c c c c c c c }
\hline
& \textbf{feminism}   & \textbf{law}            & \textbf{\begin{tabular}[c]{@{}c@{}}gender \\identity\end{tabular}} & \textbf{\begin{tabular}[c]{@{}l@{}}marrage/\\divorce\end{tabular}} & \textbf{church}     & \textbf{motherhood} & \textbf{birth control}  & \textbf{life}       & \textbf{sexuality}   & \textbf{parenting}   \\ \hline
& \textbf{Topic 0}           & \textbf{Topic 1}          & \textbf{Topic 2}    & \textbf{Topic 3}         & \textbf{Topic 4}      & \textbf{Topic 5}   & \textbf{Topic 6}   & \textbf{Topic 7}  & \textbf{Topic 8} &   \textbf{ Topic 9}               \\ \hline
1 &     women      & abortion       & gender          & marriage        & church     & mom        & humanae        & human      & sex         & parents     \\ \hline
2 &     abortion   & court          & lgbt            & divorce         & catholic   & god        & pill           & social     & sexual      & school      \\ \hline
3 &     pro        & constitutional & identity        & family          & bishops    & mary       & vitae          & ecology    & men         & children    \\ \hline
4 &     men        & circuit        & transgender     & children        & pope       & christ     & contraception  & work       & regnerus    & schools     \\ \hline
5 &     life       & federal        & ideology        & marital         & synod      & mother     & control        & revolution & cheap       & federalist  \\ \hline
6 &     feminist   & decision       & trans           & marriages       & priests    & jesus      & birth          & people     & consent     & education   \\ \hline
7 &     march      & abortions      & sex             & divorced        & francis    & like       & women          & family     & women       & students    \\ \hline
8 &     feminists  & state          & sexual          & spouses         & women      & baby       & contraceptives & moral      & desire      & transgender \\ \hline
9 &     woman      & kansas         & male            & annulment       & rome       & love       & health         & political  & porn        & child       \\ \hline
10 &    feminism   & supreme        & reality         & families        & letter     & motherhood & effects        & politics   & weinstein   & reading     \\ \hline
11 &    equality   & law            & female          & spouse          & christ     & child      & mandate        & man        & market      & kids        \\ \hline
12 &    female     & medicaid       & activists       & married         & mccarrick  & world      & fertility      & dignity    & marriage    & youth       \\ \hline
13 &    choice     & roe            & catholic        & abandoned       & vatican    & life       & iud            & person     & mating      & girl        \\ \hline
14 &    male       & constitution   & biological      & love            & bishop     & charlie    & catholic       & ecological & power       & family      \\ \hline
15 &    movement   & judge          & dysphoria       & catholics       & holy       & children   & contraceptive  & society    & pornography & gender      \\ \hline
16 &    vulnerable & case           & person          & church          & god        & light      & sanger         & nature     & fly         & confused    \\ \hline
17 &    sex        & planned        & agenda          & fidelity        & faithful   & son        & medical        & world      & good        & continue    \\ \hline
18 &    today      & parenthood     & people          & older           & faith      & little     & prevent        & liberty    & like        & boy         \\ \hline
19 &    time       & right          & orientation     & situations      & priesthood & got        & sexual         & self       & risk        & public      \\ \hline
20 &    human      & government     & report          & husband         & authority  & day        & free           & middle     & kind        & news        \\ \hline
\end{tabular}
}
\end{table}
\section{Results}\label{Sec4}
In this section, we present the results of our analysis based on the contents of ISIS propaganda materials as compared to articles from the Catholic women forum. We then present the results of emotion analysis conducted on both corpora. 
\subsection{Content Analysis}
After pre-processing the text, both corpora were analyzed for word frequencies. These word frequencies have been normalized by the number of words in each corpus. Figure~\ref{fig:word_count} shows the most common words in each of these corpora.

A comparison of common words suggests that those related to marital relationships~( husband, wife, etc.) appear in both corpora, but the religious theme of ISIS material  appears to be stronger. A stronger comparison can be made using topic modeling techniques to discover main topics of these documents. Although we used LDA, our results by using NMF outperform LDA topics, due to the nature of these corpora. Also, fewer numbers of ISIS documents might contribute to the comparatively worse performance. Therefore, we present only NMF results. Based on their coherence, we selected 10 topics for analyzing within both corporas. Table~\ref{table:isis} and Table~\ref{table:cat} show the most important words in each topic with a general label that we assigned to the topic manually. Based on the NMF output, ISIS articles that address women include topics mainly about Islam, women's role in early Islam, hijrah~(moving to another land), spousal relations, marriage, and motherhood. 

The topics generated from the Catholic women forum are clearly quite different. Some, however, exist in both contexts. More specifically, marriage/divorce, motherhood, and to some extent spousal relations appeared in both generated topics. This suggests that when addressing women in a religious context, these may be very broadly effective and appeal to the feminine audience. More importantly, suitable topic modeling methods will be able to identify these similarities no matter the size of the corpus we are working with. Although, finding the similarities/differences between topics in these two groups of articles might provide some new insights, we turn to emotional analysis to also compare the emotions evoked in the audience.

\subsection{Emotion Analysis}
\begin{figure}[!bht]
\centering
\includegraphics[width=0.66\columnwidth]{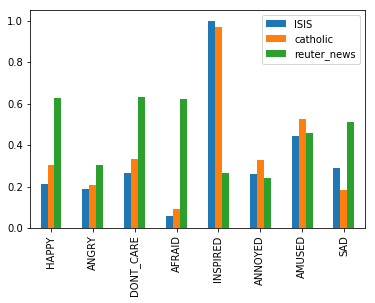}
\caption{Comparison of emotions of our both corpora along with Reuters news}\label{fig:feel_compare}
\end{figure}

We rely on Depechemood dictionaries to analyze emotions in both corpora. These dictionaries are freely available and come in multiple arrangements. We used a version that includes words with their part of speech~(POS) tags. Only words that exist in the Depechemood dictionary with the same POS tag are considered for our analysis. We aggregated the score for each word and normalized each article by emotions. To better compare the result, we added a baseline of 100 random articles from a Reuters news dataset as a non-religious general resource which is available in an NLTK python library. Figure~\ref{fig:feel_compare} shows the aggregated score for different feelings in our corpora.

\begin{figure}
\centering
\includegraphics[width=.66\columnwidth]{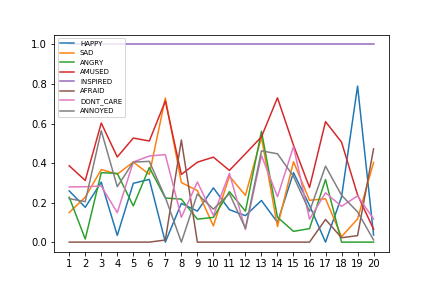}
\caption{Feeling detected in ISIS magazines~(first 7 issues belong to Dabiq and last 13 belong to Rumiyah)}\label{fig:feel_issue}
\end{figure}

Both Catholic and ISIS related materials score the highest in ``inspired'' category. Furthermore, in both cases, being afraid has the lowest score. However, this is not the case for random news material such as the Reuters corpus, which are not that inspiring and, according to this method, seems to cause more fear in their audience. We investigate these results further by looking at the most inspiring words detected in these two corpora. Table~\ref{table:ispire-words} presents 10 words that are among the most inspiring in both corpora. The comparison of the two lists indicate that the method picks very different words in each corpus to reach to the same conclusion. Also, we looked at separate articles in each of the issues of ISIS material addressing women.  Figure~\ref{fig:feel_issue} shows emotion scores in each of the 20 issues of ISIS propaganda. As demonstrated, in every separate article, this method gives the highest score to evoking inspirations in the reader. Also, in most of these issues the method scored ``being afraid'' as the lowest score in each issue.


\begin{table}[h] \addtolength{\tabcolsep}{+2pt} 
\centering
\caption{Words with highest inspiring scores}\label{table:ispire-words}
\begin{tabular}{ l c c } 
\hline
  \diagbox[width=\dimexpr \textwidth/10+12\tabcolsep\relax, height=1cm]{ Words }{Group}      & \textbf{~~~~~~~Catholic~~~~~~~}        & \textbf{~~~~~~~~~ISIS~~~~~~~~~}         \\ \hline
$Word_1$  & avarice         & uprightness  \\ \hline
$Word_2$  & perceptive      & memorization \\ \hline
$Word_3$  & educationally   & merciful     \\ \hline
$Word_4$  & stereotypically & affliction   \\ \hline
$Word_5$  & distrustful     & gentleness   \\ \hline
$Word_6$  & reverence       & masjid       \\ \hline
$Word_7$  & unbounded       & verily       \\ \hline
$Word_8$  & antichrist      & sublimity    \\ \hline
$Word_9$  & loneliness      & recompense   \\ \hline
$Word_{10}$ & feelings        & fierceness   \\ \hline
\end{tabular}
\end{table}

\section{Conclusion and Future Work}
\label{Sec5}
In this paper, we have applied natural language processing methods to ISIS propaganda materials in an attempt to understand these materials using available technologies. We also compared these texts with a non-violent religious groups'~(both focusing on women related articles) to examine possible similarities or differences in their approaches. To compare the contents, we used word frequency and topic modeling with NMF. Also, our results showed that NMF outperforms LDA due to the niche domain and relatively small number of documents. 

The results suggest that certain topics play a particularly important roles in ISIS propaganda targeting women. These relate to the role of women in early Islam, Islamic ideology, marriage/divorce, motherhood, spousal relationships, and hijrah~(moving to a new land). 

Comparing these topics with those that appeared on a Catholic women forum, it seems that both ISIS and non-violent groups use topics about motherhood, spousal relationship, and marriage/divorce when they address women. Moreover, we used Depechemood methods to analyze the emotions that these materials are likely to elicit in readers. The result of our emotion analysis suggests that both corpuses used words that aim to inspire readers while avoiding fear. However, the actual words that lead to these effects are very different in the two contexts. Overall, our findings indicate that, using proper methods, automated analysis of large bodies of textual data can provide novel insight insight into extremist propaganda that can assist the counterterrorism community.   

\bibliographystyle{bibtex/spmpsci} 
\bibliography{template.bib}

\end{document}